\documentclass[pmlr]{jmlr}

\usepackage[load-configurations=version-1]{siunitx} 

\usepackage{makecell} 

\usepackage{multirow}
\usepackage{booktabs}

\usepackage{longtable}

\usepackage[capitalise]{cleveref}
\usepackage{siunitx}

\usepackage{rotating}

\usepackage{multirow} 

\usepackage{tcolorbox} 

\usepackage{enumitem}

\usepackage{caption}

\usepackage{listings}
\usepackage{xcolor}
\usepackage{graphicx}

\definecolor{keywords}{RGB}{127,159,191}
\definecolor{comments}{RGB}{147,112,219}
\definecolor{red}{RGB}{220,50,47}
\definecolor{green}{RGB}{138,226,52}
\definecolor{dkgreen}{rgb}{0,0.6,0}
\definecolor{gray}{rgb}{0.5,0.5,0.5}
\definecolor{mauve}{rgb}{0.58,0,0.82}
\definecolor{mygray}{rgb}{0.9,0.9,0.9}
\definecolor{LightGray}{gray}{0.95}

\usepackage[switch]{lineno}

\newcommand{\equal}[1]{{\hypersetup{linkcolor=black}\thanks{#1}}}

\usepackage{soul}
\usepackage{ulem}

\colorlet{soulgreen}{green!30}
\sethlcolor{soulgreen}%

\newcolumntype{R}{>{$}r<{$}}

\theorembodyfont{\upshape}
\theoremheaderfont{\scshape}
\theorempostheader{:}
\theoremsep{\newline}

\usepackage{pifont}
\newcommand{\xmark}{\ding{55}}
\jmlrvolume{} 
\jmlryear{} 
\jmlrworkshop{} 

\title[]{
Iterative Learning of Computable Phenotypes for Treatment Resistant Hypertension using Large Language Models
}

\author{%
\Name{Guilherme Seidyo Imai Aldeia} \Email{guilherme.aldeia@ufabc.edu.br} \\
\addr{Federal University of ABC, Santo Andr\'{e}, S\~{a}o Paulo, Brazil}
\AND
\Name{Daniel S. Herman}\equal{These authors contributed equally} \Email{Daniel.herman2@pennmedicine.upenn.edu}\\
\addr{Department of Pathology and Laboratory Medicine, University of Pennsylvania, Philadelphia, PA, USA}
\AND
\Name{William G. {La~Cava}}\footnotemark[1] \Email{william.lacava@childrens.harvard.edu}\\
\addr{Computational Health Informatics Program, Boston Children's Hospital, Harvard Medical School, Boston, MA, USA}
}

\begin{document}

\maketitle

\begin{abstract}
Large language models (LLMs) have demonstrated remarkable capabilities for medical question answering and programming, but their potential for generating interpretable computable phenotypes (CPs) is under-explored.  
In this work, we investigate whether LLMs can generate accurate and concise CPs for six clinical phenotypes of varying complexity, which could be leveraged to enable scalable clinical decision support to improve care for patients with hypertension.
In addition to evaluating zero-short performance, we propose and test a \textit{synthesize, execute, debug, instruct} strategy that uses LLMs to generate and iteratively refine CPs using data-driven feedback.
Our results show that LLMs, coupled with iterative learning, can generate interpretable and reasonably accurate programs that approach the performance of state-of-the-art ML methods while requiring significantly fewer training examples. 
\end{abstract}

\section{Introduction}
\label{sec:intro}

Recent advances in neural network architectures, particularly transformers \citep{NIPS2017_3f5ee243}, have led to groundbreaking innovations in large language models (LLM), enabling commercial products such as ChatGPT \citep{openai_chatgpt} and Claude \citep{anthropic2024claude3}. As a result, LLMs have been widely applied across various domains, demonstrating their effectiveness in solving domain-specific problems, including in coding \citep{jiang2024surveylargelanguagemodels}, law \citep{LAI2024181}, and healthcare \citep{zhou2023survey, Thirunavukarasu2023}.

Given the complexity of LLMs and other artificial intelligence (AI) models, the robustness and intelligibility of their responses requires scrutiny. The interpretability of the recommendations of AI tools is an important factor considered by regulatory agencies, such as the US FDA \citep{FDA} and the European AI Office \citep{europeanunionregulations}, in assessing the risks of deploying AI-targeted decision support in healthcare.
Post-hoc explanation methods~\citep{Lundberg2017}, including recent methods applying representation engineering \citep{zou2023representation} and sparse auto-encoders \citep{ng2011sparse}, are fundamentally limited in that they cannot fully explain models' behavior ~\citep{Ghassemi2021,Rudin2019}. 
Additionally, LLM models are prone to \textit{hallucinations} (\textit{i.e.}, generating fabricated information) \citep{ouyang2022training,singh2024rethinking}, and can show bias across demographic subgroups \citep{chen2024crosscareassessinghealthcareimplications}.

While previous research has explored LLMs for natural language processing (NLP) tasks, an overlooked question is whether they can generate useful \emph{computable phenotypes} (CPs) \citep{bandaAdvancesElectronicPhenotyping2018a}.
A CP is an algorithmic construct that identifies observable traits from patient electronic health record data \citep{he2023trends}, enabling the identification of patients with a shared condition of interest \citep{tasker_why_2017}.
CPs are desirable for decision support because they are machine-executable and can also be both intuitively interpretable to clinicians and can precisely identify specific patient populations~\citep{moDesiderataComputableRepresentations2015a}. 
However, conventional manual approaches to the construction of CPs require considerable time and effort from clinical experts and data analysts, and as such, they do not scale well across phenotypes and struggle to adapt to differences across clinical practices or changes over time.
In fact, \citet{he2023trends} reviewed CP generation literature, finding that 40\% of the CP does not use ML --- and when CP generation relies on ML, authors often choose models such as SVM and random forests, and perform feature selection manually, by choosing features related to previous studies, or guidelines, but rarely use automated generation processes.
As LLMs are trained on medical literature and structured datasets \citep{liu2024datasets} and perform well on medical Q\&A tasks, they likely encode sufficient information regarding the relationships among clinical concepts to connect clinical concepts provided in prompts to available EHR data elements, enabling the composition of CPs.
As opposed to the direct usage of LLMs for making clinical predictions ~\citep{jiangHealthSystemscaleLanguage2023}, using LLMs to compose CPs may yield efficient, standalone models that can be transparently assessed for performance and intelligibility.

Computable phenotyping is inherently a \textit{program synthesis} problem, for which the goal is to solve a computer programming problem autonomously. 
The task of generating CPs as a program synthesis problem was studied using supervised machine learning (ML) methods, such as symbolic regression (SR), which simultaneously optimizes the model structure and parameters to produce simple and interpretable models~\citep{lacavaFlexibleSymbolicRegression2023}. 
Recent work has shown the potential for LLMs to tackle general program synthesis~\citep{liventsevFullyAutonomousProgramming2023}, especially when used as components in an iterative process including the steps of synthesizing, executing, and debugging ~\citep{guptaSynthesizeExecuteDebug2020}.

As a new technology, it is largely unknown how to best apply current LLMs for CP development. 
To that end, this paper aims to address the following research questions:
\begin{enumerate}
    \item  Can LLMs generate clinically meaningful CPs for hypertension-related conditions? If so, how much detail does the prompt require? 
    \item How accurate and concise are LLM-generated models compared to those produced by interpretable ML methods?
    \item Can an iterative refinement improve LLM-generated CPs?
\end{enumerate}

We investigate the capability of LLMs to generate CPs for three phenotypes of increasing complexity: hypertension (HTN), hypertension with unexplained hypokalemia (HTN-HypoK), and apparent treatment-resistant hypertension (aTRH). Because LLMs are trained on large corpuses of code, we decide to explore their ability to compose CPs as simple Python programs. 
Our approach differs from other CP generation ones by leveraging natural language descriptions of the phenotypes to automatically guide CP construction, using less data to perform iterative optimization. We acknowledge LLMs are speculative, but as an intermediate step, we can focus on the final CP for using in clinical practice.

We targeted hypertension and these sub-phenotypes because of the substantial opportunity for leveraging such CPs for clinical decision support. 
One critical need is to improve screening for primary aldosteronism, which is recommended for patients with HTN-HypoK or aTRH by specialty society guidelines ~\citep{funder_management_2016}. 
Primary aldosteronism causes cardiovascular disease at rates that are even higher than other causes of hypertension ~\citep{milliez_evidence_2005}. 
It is thought to affect up to $\approx1\%$ of US adults and is treatable, and sometimes curable, when identified but it is only diagnosed in fewer than $10\%$ of affected patients ~\citep{kayser_study_2016, funder_primary_2016, cohen_testing_2020, reincke_observational_2012, lin_adrenalectomy_2012, catena_long-term_2007}.  
Pilot studies have demonstrated that targeted decision support can increase screening of eligible patients from $2\%$ to at least $16\%$ ~\citep{passman_active_2025}, but it is difficult to manually adapt manually curated CPs from one clinical practice to another. 
By developing methodologies for automatically generating CPs, we can create a largely automated pipeline for adapting CPs across different localities or cohorts and adapting them over time to maintain performance.

We evaluate various aspects of CP generation, including:
\textit{i)} different LLMs,
\textit{ii)} different degrees of detail in the specification of the desired CP,
\textit{iii)} the number of features provided to the LLM, and
\textit{iv)} an iterative \textit{synthesize, execute, debug, and instruct} (SEDI) strategy, wherein misclassified patients are provided as feedback to the LLM to refine the CP.

Our experiments are conducted using pre-processed EHR data. We compare LLM-generated models to interpretable ML methods, including decision trees, logistic regression, and a symbolic regression framework called FEAT \citep{lacavaLearningConciseRepresentations2019}.

Our results showed that the LLMs generated concise CPs for all phenotypes analyzed. As expected, LLM-generated CPs were more accurate when the prompt included a more detailed description of the desired phenotype. 
The SEDI strategy improved performance, even when prompts did not include detailed phenotype definitions. 
To some extent, GPT-based models under-performed relative to the best supervised ML approaches in terms of cross-validated area under the precision-recall curve (AUPRC). 
However, the performance of the best LLM-generated CP (\texttt{gpt-4o+SEDI}) exhibited similar AUPRC and AUROC to a recently published, ML-based CP for predicting aTRH~\citep{lacavaFlexibleSymbolicRegression2023} in held-out testing data. 
When performing parameter optimization on the final model, we could further increase its AUPRC performance in held-out testing data, outperforming the ML-based CP.

\subsection*{Generalizable Insights about Machine Learning in the Context of Healthcare}


Computable phenotypes are useful tools for clinicians whose development traditionally demands a large amount of clinician time spent in chart review and model refinement. 
The expensive nature of the gold-standard labeling process for this task presents a general challenge for ML solutions. 
Our results suggest that LLMs can be leveraged to automate both the generation of computable phenotypes and the review and refinement process using a smaller amount of expert-curated samples than is typically required to train ML models. 
We further find that, by using LLMs to generate CPs rather than black-box predictions, the model behavior becomes interpretable.  

\section{Related work}
\label{sec:related}

In healthcare, LLMs have been employed as NLP tools for tasks such as extracting concepts from clinical notes for postpartum hemorrhage patients without task-specific training (zero-shot) \citep{Alsentzer2023}, making meaningful inferences from physiological and behavioral time-series data by providing a few relevant examples in the prompt (few-shot) \citep{liu2023large}, and extracting logical conditions of inclusion/exclusion for phenotypes using concept codes \citep{Tekumalla2024}.
LLMs are trained on vast amounts of data \citep{NEURIPS2022_c1e2faff}, and can learn complex linguistic structures and rich representations of real-world information, thereby achieving remarkable performance in zero-shot or few-shot tasks.
\citet{jiangHealthSystemscaleLanguage2023} described LLMs as ``all-purpose prediction engines'' for health systems due to their ability to perform well across five seemingly disparate tasks (30-day all-cause readmission prediction, in-hospital mortality prediction, comorbidity index prediction, length of stay prediction, and insurance denial prediction).
The performance of LLMs on medical exam question-answering has become a \textit{de facto} benchmark to track AI model advances~\citep{jinWhatDiseaseDoes2020a,singhalLargeLanguageModels2023a}, although such benchmarking has recently been called into question given its specious relationship to usefulness in real-world tasks~\citep{rajiItsTimeBench2025}.

More detailed investigations into specific clinical applications of LLMs for EHR data, reviewed by~\citet{wornowShakyFoundationsLarge2023}, suggest that most meaningful clinical applications of LLMs, outside of high-level prediction tasks, are under-explored. 

LLMs were recently evaluated for their ability to draft CPs in the form of SQL queries for three phenotypes: type 2 diabetes mellitus, dementia, and hypothyroidism~\citep{yanLargeLanguageModels2024}. Another promising evaluation was done by \citet{Tekumalla2024} of usefulness and reliability in retrieving phenotype concept codes was done by comparing LLM-generated phenotypes with OHDSI phenotype library \citep{banda2017electronic} and HDRUK phenotype library \citep{HRDUK2025} by domain-experts.

Beyond clinical applications, LLMs have also been employed to generate concepts for discriminating input texts, then aggregating the concept scores into a white-box prediction model \citep{ludan2024interpretablebydesigntextunderstandingiteratively}.
With a focus on medical calculations, MedCalc-Bench \citep{khandekar2024medcalcbenchevaluatinglargelanguage} evaluated the task of predicting common medical scores based on textual descriptions of patient records.

While these studies show LLMs in healthcare applications and computable phenotyping, to the best of our knowledge, no prior work directly aligns with our approach.
Existing research has primarily focused on using LLMs for prediction tasks, medical question-answering, and phenotype retrieval, largely in the zero-shot context. 
In contrast, our study provides a detailed investigation into generating CPs with LLMs, including through automated refinement, in comparison to expert and ML-based solutions.

\section{Methods}
\label{sec:methods}

LLMs generate text by predicting the next token in a sequence, using as context the previous tokens. A token is a fundamental unit of text that can be a word or sub-word, encoded in a numerical embedding. 
The novelty introduced by transformers is the efficient processing of long-range token-token dependencies.
By default, LLMs select the most probable next token given the specified preceding context.
Some parameters, such as \textit{temperature} (which perturbs the probability space when generating the next token) \citep{grubisic2024prioritysamplinglargelanguage} or \textit{nucleus sampling} (which restricts token choices to within a cumulative probability threshold) \citep{ravfogel2023conformal} adjust the variability in selection of the next token.

To enhance their ability to follow structured instructions, LLMs can be fine-tuned to the \textit{instruct} task \citep{wei2022finetuned}, using datasets containing instruction-response pairs.
In this setup, prompts are categorized into \textit{system prompts} (guiding overall model behavior) and \textit{user prompts} (task-related queries).
Instruct models have no memory, so the entire conversation is used as context when generating a new response.

\subsection{LLM-Guided CP Generation}

\begin{figure*}[htb]
    \centering
    \includegraphics[width=\linewidth,trim={1.25cm 0.25cm 0.5cm 0}, clip=true]{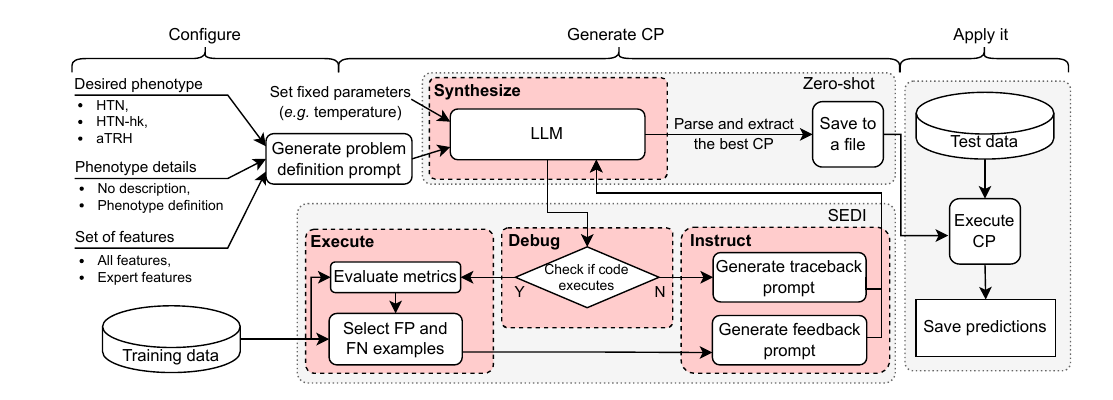}
    \caption{Overview of our proposed method for generating computable phenotypes with LLMs. A prompt is constructed by selecting a phenotype, specifying the level of description detail, and choosing the set of available features. In the zero-shot approach, the LLM generates a Python function to estimate phenotype probability using the listed features. With the SEDI strategy, a loop performs the following iteration: the CP is applied to the training dataset and the LLM is informed regarding the overall performance metrics and provided with false positive (FP) and false negative (FN) examples. Finally, after saving the CP to a file, it can be used to predict probabilities for held-out data as a standalone function.}
    \label{fig:pipeline-llms}
\end{figure*}

\cref{fig:pipeline-llms} illustrates our approach for leveraging LLMs to construct CPs. 
We explore three target phenotypes, specify these phenotypes in the prompt using two levels of detail, and provide either the full set of features or a smaller, pre-selected set of features.

We explore two modes for prompt composition:

\paragraph{Zero-shot prompts} The LLM generates a Python function that predicts phenotype probabilities based on the available features without receiving feedback.

\paragraph{SEDI prompts} The process follows a synthesize-execute-debug-instruct (SEDI) loop \citep{guptaSynthesizeExecuteDebug2020} iteratively receiving feedback regarding the CP's performance on training dataset. 
If the CP fails to execute, the LLM receives a message containing the error traceback (Debug). 
If the CP is successful, the LLM receives performance metrics, as well as example false positive (FP) and example false negative (FN) cases, and is instructed to refine its phenotype definition to improve the program's performance (Instruct). 
When providing FP and FN cases, information about the features being currently used by the phenotype, and additionally randomly sampled features are provided.
We set to $10$ false positive and $10$ false negative examples per iteration, capped at $10$ iterations --- using at most $200$ samples. In practice, samples may be resampled, and some iterations may contain fewer false positive or false negative examples, with $200$ serving as a strict upper bound.

The SEDI prompts relies on giving examples for the model to perform small adjustments with supervised feedback, as including examples of inputs and desired outputs can improve overall performance \citep{murr2023testing}. 
Our integration of the SEDI strategy significantly reduces reliance on labeled data compared to traditional supervised machine learning methods. 
In our approach, we retain the history of all models generated during the iterative phase, and at the end select the model that performs the best on training data as the final CP.

The \textit{problem definition} prompts are presented below. The additional SEDI prompts are detailed in \appendixref{apd:sedi-prompts}.

\paragraph{System prompt} You are an AI assistant that generates Python code based on a plain-text description of a function's purpose. You will receive a statement describing what the function should do. Your response must contain only a Python function, with no comments or explanations, that strictly follows the given description.

\paragraph{User prompt} Please create a Python function named `predict\_hypertension' that takes a pandas DataFrame named `df' as input. The function should assess whether each patient (represented as rows) has evidence of \textless{}\texttt{phenotype} description\textgreater{}. The function must return an array of floats representing the probability for each row. The available columns and their meanings are provided as key value pairs in the following dictionary: `\textless{}\texttt{variable dict}\textgreater{}'. You may only use the features whose names appear in this dictionary.

\section{Cohort} 

Our experiments were conducted using EHR data from $1200$ patients who received longitudinal primary care in the University of Pennsylvania Healthcare System (UPHS) and had chart review. 
The study data was previously reported in \citet{lacavaFlexibleSymbolicRegression2023}. 
Subject inclusion criteria were: (a) at least five outpatient visits in at least three separate years between $2007$ and $2017$, (b) at least two encounters at a single primary care practice sites, and (c) age $18$ years or older.
\cref{tab:statistics-data} describes the population comprising the data, including distribution of the three diagnoses we study here.

\begin{table}[htb]
    \footnotesize
    \caption{
    Demographic and diagnosis statistics of the study population.
    Values are counts (\%) unless otherwise noted.
    }
    \label{tab:statistics-data}
    \centering
    \renewcommand\arraystretch{1.1}
\newcommand{\colw}{4em}
\begin{tabular}{@{}llllc@{}}
\toprule\midrule
 &  & & \multicolumn{2}{c}{Grouped by aTRH} \\
\cline{4-5}
 &  & Overall & False & True \\
\midrule
n &  & 1199 & 1023 & 176 \\
\cline{1-5}
\multicolumn{2}{l}{Age, mean (SD)} & 57.2 (18.5) & 55.0 (18.2) & 70.4 (14.0) \\
\cline{1-5}
\multirow[t]{3}{\colw}{Race} & Black & 338 (28.2) & 251 (24.5) & 87 (49.4) \\
 & Other & 108 (9.0) & 100 (9.8) & 8 (4.5) \\
 & White & 753 (62.8) & 672 (65.7) & 81 (46.0) \\
\cline{1-5}
\multirow[t]{2}{\colw}{Sex} & F & 737 (61.5) & 631 (61.7) & 106 (60.2) \\
 & M & 462 (38.5) & 392 (38.3) & 70 (39.8) \\
\cline{1-5}
\multirow[t]{2}{\colw}{HTN} & False & 591 (49.3) & 591 (57.8) & 0 (0) \\
 & True & 608 (50.7) & 432 (42.2) & 176 (100) \\
\cline{1-5}
\multirow[t]{2}{\colw}{HTN HypoK} & False & 1027 (85.7) & 924 (90.3) & 103 (58.5) \\
 & True & 172 (14.3) & 99 (9.7) & 73 (41.5) \\
\midrule
\bottomrule
\end{tabular}

\end{table}

The data contains a total of $331$ features extracted from EHR data. 
The features include demographic and clinical features such as age, sex, race, hospital proximity, weight, BMI, blood pressure, occurrences of elevated BP (systolc/diastolic $\geq 140/90$ mmHg) during encounters.
Longitudinal features were aggregated as minimum, maximum, median, standard deviation, and skewness.

Laboratory results from $34$ common tests (\textit{e.g.}, metabolic panel, blood count, lipids, TSH, HbA1c) were aggregated using quartiles and medians. 
Diagnosis codes for hypertension and related comorbidities were summarized as annual medians and totals. 
Medication prescriptions were summarized as the number of days prescribed for each antihypertensive class and the count of encounters while prescribed $1$, $2$, $3$, or $4$ or more anti-hypertensive medications, described as sum, median, standard deviation, and skewness.
Clinical notes were scanned for mentions of hypertension using regular expressions.

Features with physiologically implausible values, low variance ($<0.05$), or sparse data ($<5\%$ non-zero counts) were excluded.
Missing values were imputed using the median. A detailed description is available in \appendixref{apd:all-feature-description}.

\subsection{Target computable phenotypes}

Three clinical phenotypes were targeted, in order of increasing complexity: hypertension (HTN), hypertension with hypokalemia (HTN-HypoK), and apparent treatment-resistant hypertension (aTRH). 
For each phenotype, two labeling methods were used: ``diagnosis'' (Dx) based on expert chart reviews and ``heuristic'', an expert-designed clinical protocol refined iteratively using training data to work as a simpler version of the phenotypic labels.
Heuristics were initially developed on the basis of clinical data expertise and iteratively refined based on evaluation in subsets of the training data in this study.
By evaluating performance on both heuristics and diagnosis labels, we can assess how well the LLM captures simpler heuristics and, subsequently, how well it represents more complex phenotypes.

The simple HTN heuristic consisted of the presence of two or more diagnosis codes for hypertension\footnote{The data were recorded, pre-processed, and labeled, under the ICD-9 classification, which was in use at the time.}. 
For HTN-HypoK, the heuristic was the presence of the HTN heuristic plus two of one of the following: diagnosis codes for hypokalemia, outpatient encounters with low blood potassium results, or ambulatory prescriptions for an oral potassium supplement. 
For aTRH, we refined a previously reported CP to label patients. 
It consisted of inclusion criteria: (a) with documentation of at least two outpatient encounters with elevated blood pressure while on antihypertensive medications from 3 distinct classes or (b) prescribed four or more distinct classes of antihypertensive medications. 
The exclusion criteria for aTRH included diagnosis codes for heart failure, heart transplant, or moderate to severe chronic kidney disease prior to meeting the above inclusion criteria.

For diagnosis labels ascertained by chart review, a study physician reviewed clinical charts and classified subjects with respect to three phenotypes. The classification was based on JNC7 Guidelines on Prevention, Detection, Evaluation, and Treatment of High Blood Pressure \citep{chobanian2003seventh}.

\subsection{Prompt construction with phenotype and data description}

We start by describing our definitions for each hypertension phenotype, used when constructing prompts with detailed phenotype descriptions. They were designed to provide precise context to the model, when cross-referenced with the provided dictionary for available EHR features.

\begin{description} \small
    \item[Hypertension:] 2 or more hypertension Dx codes;
    \item[Hypertension with hypokalemia:] 2 or more hypertension Dx codes and either 2 or more low potassium test results, 2 or more potassium supplementation prescriptions, or 2 or more hypokalemia diagnosis codes;
    \item[Treatment resistant hypertension:] 2 or more high blood pressure measurements while prescribed 3 or more hypertension medications or 2 or more encounters while prescribed 4 or more hypertension medications
\end{description}

We identified a small feature set sufficient for either precisely building or approximating the heuristics definitions, described in \cref{tab:features-in-phenotype-definition}. 
To ensure the LLMs appropriately interpret the features' derivations, we provide precise feature descriptions. 
When not using the subset of features from \cref{tab:features-in-phenotype-definition}, then the model is prompted with all features and their respective descriptions, potentially allowing it to capture more complex dependencies but creating a much larger prompt.

\begin{table*}[htb]
    \caption{
    The ``expert" features used to define the phenotypes in the expert-curated heuristics. 
    Under the `expert features' experiment setting, rather than receiving a data dictionary of all available features ($n$=311), the LLM receives this focused set.
    }
    \label{tab:features-in-phenotype-definition}
    \centering
    \footnotesize
    \begin{tabular}{@{}rl@{}}
    \toprule\midrule
    Feature name & Feature description \\ \midrule
    mean/median\_systolic & Mean/median of systolic blood pressure (SBP) measured \\
    mean/median\_diastolic & Mean/median of diastolic blood pressure (DBP) measured \\
    bp\_n & Total number of blood pressure (BP) measurements \\
    high\_bp\_n &  \begin{tabular}[c]{@{}l@{}}Number of high blood pressure measurements \\ (SBP \textgreater{}= 140 or DBP \textgreater{}= 90)\end{tabular} \\
    high\_BP\_during\_htn\_meds\_X & \begin{tabular}[c]{@{}l@{}}Number of high BP measurements (SBP \textgreater{}=140 or DBP \textgreater{}=90)\\ while prescribed X hypertension medications (X=1, 2, 3)\end{tabular} \\
    high\_BP\_during\_htn\_meds\_4\_plus & \begin{tabular}[c]{@{}l@{}}Number of high BP measurements (SBP \textgreater{}=140 or DBP \textgreater{}=90)\\ while prescribed four or more hypertension medications\end{tabular} \\
    sum\_enc\_during\_htn\_meds\_4\_plus & Total encounters while prescribed 4 or more hypertension medications \\
    low\_K\_N & Total number of low potassium test results \\
    test\_K\_N & Total number of potassium test results \\
    Med\_Potassium\_N & Total number of potassium supplement prescriptions \\
    Dx\_HypoK\_N & Total number of hypokalemia diagnoses \\
    re\_htn\_sum & Sum of regex counts for hypertension in clinical notes \\
    \midrule\bottomrule
    \end{tabular}
\end{table*}

By using the prompt presented in \sectionref{sec:methods} under ``User prompt'', we replace \textless{}\texttt{phenotype}\textgreater{} by the description. 
A simple prompt will contain only the name of the phenotype (i.e. ``hypertension"), while the rich prompt is formatted as follows:  \textless{}\texttt{phenotype}\textgreater{}, which we will define as \textless{}\texttt{description of the heuristic}\textgreater{}.

\section{Experiments}
\label{sec:experiments}

We evaluated LLMs' capabilities in generating CPs for hypertension and subphenotypes thereof, and evaluated their performance on both the expert-defined heuristics and chart-reviewed patient diagnoses.
The biggest knowledge gaps include understanding how prompts influence phenotype generation, whether adding more features improves accuracy or unnecessarily increases complexity, and how LLM-composition compares with existing ML methods for the same task.

In our first set of experiments, we perform an extensive evaluation of different settings for the LLMs with each phenotype.
We investigate how different LLM models respond to prompt richness and feature quantity, and whether the SEDI loop improves performance over ten iterations.
\cref{tab:llm-models-used} reports the LLMs evaluated in this study.
We set the temperature to $0.5$ and nucleus sampling (\texttt{top-p}) to $1.0$, based on preliminary experiments, to allow some variability while keeping the generated programs concise.

\begin{table}
\footnotesize
    \caption{
    LLM models considered in our experiments, accessed via the OpenAI API. 
    }
    \label{tab:llm-models-used}
    \centering
    \begin{tabular}{@{}llll@{}}
    \toprule\midrule
    Model & Checkpoint date (yy-mm-dd) & Knowledge cut-off date (yy-mm) & Max tokens \\
    \midrule
    \texttt{gpt-3.5-turbo} & 24/04/09 & 21/10 & 16K \\
    \texttt{gpt-4o-mini} & 24/7/18 & 23/10 & 128K  \\
    \texttt{gpt-4o} & 24/08/06  & 23/10 & 128K \\
    \midrule\bottomrule
    \end{tabular}
\end{table}

Our second set of experiments focuses on comparing the best setting for prompt richness and feature set with other interpretable ML methods, namely decision trees (DT), logistic regression with L1 regularization (LR L1), random forests (RF), and a symbolic regression algorithm named Feature Engineering Automation Tool (FEAT)~\citep{lacavaLearningConciseRepresentations2019}.
Although random forests are not inherently interpretable, they are powerful models and serve as a competitive accuracy benchmark for this task.

Methods were compared using $5$-fold cross-validation (CV) over $75\%$ (n=$899$) of the study subjects. 
Each fold was tested with ten different random seeds, as some methods (\textit{e.g.}, RF, LLMs, FEAT) are stochastic.
Non-LLM based method underwent additional hyperparameter tuning on each fold (\textit{i.e.}, nested CV), using the hyperparameters in \cref{tab:other-ml-methods} in \appendixref{apd:hyperparameters}.

We first compared the cross-validation performance across all $50$ runs per phenotype and modeling strategy.
Next, we focused on the most complex phenotype, aTRH, and selected from these runs the LLM-based CP with the best cross-validation performance and evaluated it on a held-out test set of $300$ patients. The selected CPs went through an additional parameter optimization step aiming to maximize AUPRC, and the tuned version was also included in the comparisons.

\paragraph{Metrics}

To evaluate the performance of the LLM-generated CPs and ML models, we report the area under the precision-recall curve (AUPRC) and the area under the receiver operating characteristic curve (AUROC).

In addition to manually interpreting a small set of generated CPs, we measured the size of each model as a scalable proxy for potential interpretability. 
In short, we approximate this model complexity by counting the number of computational components each model contains.
DT model size is measured as the total number of nodes of the tree. 
RF is measured as the sum of number of nodes for every tree.
LR L1 is measured by the number of features and arithmetic operations needed to build a linear combination of features with non-zero coefficients.
FEAT is measured as the number of operations, constants, and features used in the mathematical function.
To estimate the size of LLM-generated CPs, we first load the generated function into Python's abstract syntax tree library (\texttt{ast}) and then count the number of nodes in progam tree. 
Although imperfect, this size comparison gives us a valuable, if rough, sense of the potential interpretability of models returned by different methods.

\section{Results}
\label{sec:results}

\begin{figure*}[htb]
    \centering
    \includegraphics[width=\linewidth]{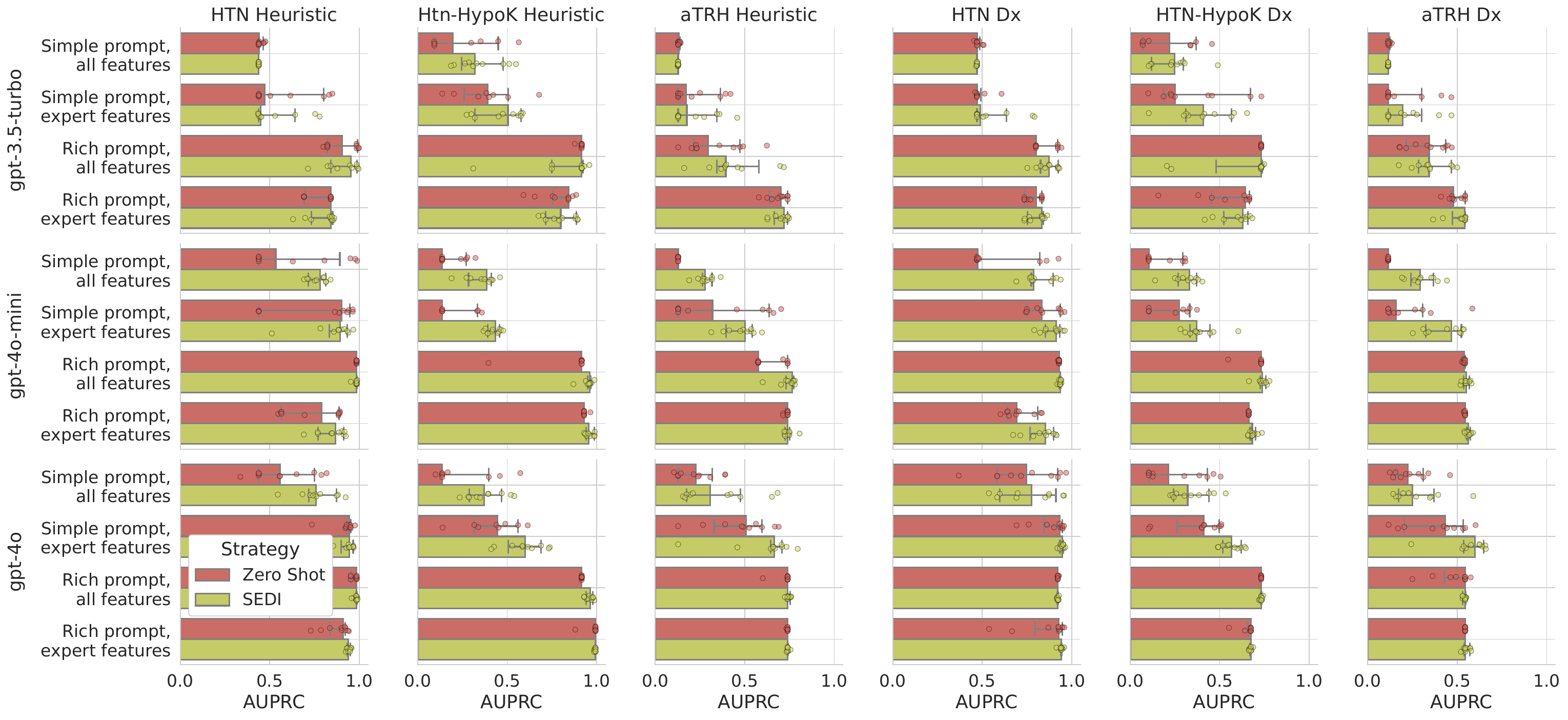}
    \caption{
        Average AUPRC on each held-out fold of the LLM-generated CPs across different prompting approaches. 
        Prompts were varied to exclude/include detailed descriptions of the phenotype (\textit{i.e.}, Simple, Rich) and include the entire set of features or only expert-selected features used in the heuristic (\textit{i.e.}, all features, expert features). Whiskers depict the $95\%$ confidence interval across different seeds.
        }
    \label{fig:ablation}
\end{figure*}

\cref{fig:ablation} reports the cross-validated AUPRC for the settings study. 
We compare the performance of each LLM (rows) in generating CPs for the six different phenotypes (columns).
For each plot on the grid, we sort in the $y$ axis each combination of the level of detail provided for the phenotype and the provided feature set, in order of increasing design complexity. 
For each combination, the bars represent the different prompting strategies.
A boxplot version of this result is presented in \Cref{apd:boxplot} to better visualize the distribution.

\begin{figure*}[htb]
    \centering
    \includegraphics[width=\linewidth]{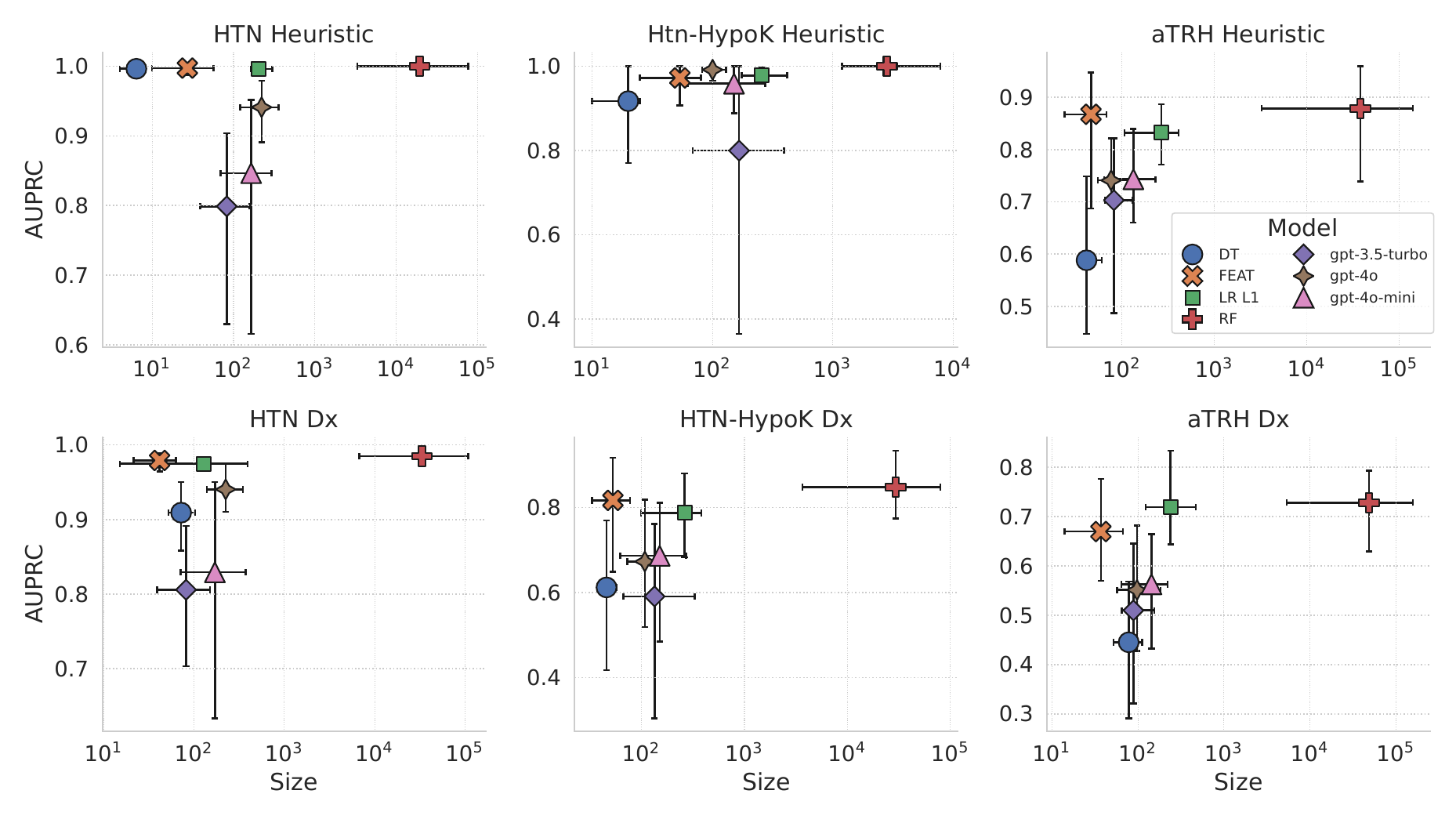}
    \caption{Trade-off between validation AUPRC (higher is better) and model size (smaller is better) for all phenotypes, using the SEDI strategy, rich prompts, and expert features. The error bars denote the $95\%$ confidence intervals.}
    \label{fig:pareto-plots}
\end{figure*}

In subsequent comparisons to other ML methods, for simplicity, we compare to LLM results using the SEDI strategy, as it appeared on average to be associated with the best overall performance for all LLM models.
As a proxy for interpretability, we measure and plot the model sizes against the cross-validation AUPRC using the best llm settings and the SEDI strategy for all phenotypes in \cref{fig:pareto-plots}, also including the models found by other ML methods.
In addition, \Cref{apd:auprc-all-methods} compares the distribution of validation AUPRC for aTRH across different LLM-generated CPs and interpretable machine learning methods.

Finally, we compared the performance of the best GPT-generated CP for aTRH to the FEAT CP reported in prior work, using 300 held-out test subjects. A single final model was selected for \texttt{gpt-4o} variants as follows: the best GPT-generated CPs were identified by selecting, from all CPs generated across cross-validation folds and iterations, the one with the highest performance.

We compare the performance of FEAT and \texttt{gpt-4o}-generated models at different levels of prompt richness and expert feature inclusion for aTRH in \cref{tab:held-out-test-metrics}.
The validation columns reports the cross-validation performance which also corresponds to the maximum observed performance on the train-validation data. The held-out column reports the performance on exclusive data with different prevalence.
Additional parameter tuning was performed using a black-box optimization framework on the LLM-generated function generated with the SEDI strategy (See details in \cref{apd:final-model-tuning}), as it iteratively refines the program, to further assess whether the generated CPs holds reasonable parameters.

\begin{table*}[htb]
    \caption{Comparison between FEAT and best LLM-generated models from cross-validation for different levels of prompt richness and expert features for the most complex phenotype aTRH Dx. We performed parameter tuning of the final models with a black-box optimization framework. Values in parentheses represent $90\%$ confidence intervals. Highlight denotes the best-performing method across all the rows and those whose confidence intervals overlap with it. Validation partition prevalence is $0.11$. Held-out partition prevalence is $0.24$.
    }
    \label{tab:held-out-test-metrics}
    \footnotesize
    \centering
    \begin{tabular}{lccccccc}
        \toprule\midrule
        \multirow{2}{*}{\begin{tabular}[c]{@{}c@{}}Model\end{tabular}} & \multirow{2}{*}{Strategy} & \multirow{2}{*}{\begin{tabular}[c]{@{}c@{}}Rich\\prompt\end{tabular}} & \multirow{2}{*}{\begin{tabular}[c]{@{}c@{}}Expert\\features\end{tabular}} & \multirow{2}{*}{\begin{tabular}[c]{@{}c@{}}Param.\\tuning\end{tabular}} & \multirow{2}{*}{\begin{tabular}[c]{@{}c@{}}Model\\size\end{tabular}} 
        & \multicolumn{2}{c}{Held-out test} \\
        \cmidrule(lr){7-8}
        & & & & & & AUROC & AUPRC \\
        \midrule
        FEAT & - & - & \xmark & \checkmark & $\mathbf{44}$ & $\mathbf{0.94\ (0.91 - 0.96)}$ & $0.80\ (0.71 - 0.87)$ \\ \midrule
        RF & - & - & \xmark & - & $5539$ & $\mathbf{0.96\ (0.95 - 0.98)}$ & $\mathbf{0.90\ (0.84 - 0.94)}$ \\ \midrule
        \multirow[c]{7}{*}{GPT-4o} & \multirow[c]{2}{*}{Zero-Shot} & \xmark & \xmark & \xmark & 68 & $0.86\ (0.86 - 0.86)$ & $0.61\ (0.61 - 0.61)$ \\
                                   & & \checkmark & \checkmark & \xmark & 8 & $0.93\ (0.90 - 0.95)$ & $0.71\ (0.64 - 0.79)$ \\
        \cmidrule{2-8}
        & \multirow[c]{5}{*}{SEDI} & \xmark & \xmark & \xmark & $57$ & $\mathbf{0.95\ (0.93 - 0.97)}$ & $\mathbf{0.86\ (0.79 - 0.91)}$ \\
                                   &  & \checkmark & \checkmark & \xmark & $59$ & $\mathbf{0.95\ (0.92 - 0.96)}$ & $0.79\ (0.72 - 0.87)$ \\
        \cmidrule{3-8}
                                   & & \xmark & \xmark & \checkmark  & $57$ & $0.90\ (0.86 - 0.92)$ & $0.74\ (0.65 - 0.82)$ \\
                                   & & \checkmark & \checkmark & \checkmark & $59$ & $\mathbf{0.94\ (0.91 - 0.96)}$ & $\mathbf{0.85\ (0.78 - 0.91)}$ \\ 
                                 
        \midrule\bottomrule
    \end{tabular}
\end{table*}

The \texttt{gpt-4o+SEDI} model with rich prompt and expert features (\cref{fig:aTRH-final-model-gpt-4o}) achieves an average AUPRC of $0.79$ and an AUROC of $0.90$, with overlapping confidence intervals when comparing to FEAT.
The parameter optimization appeared to increase the \texttt{gpt-4o+SEDI} final model's performance to AUPRC of $0.85$ and AUROC of $0.94$.
The results show that the FEAT-trained model (\cref{fig:aTRH-final-model-feat}) achieves an AUPRC of $0.80$ and an AUROC of $0.94$; overlapping confidence intervals suggest the \texttt{gpt-4o} methods are as good as the prior CP on this cohort.
We also observe improvements on AUPRC but not on AUROC between \texttt{gpt-4o+SEDI} without parameter optimization and it's correspondent \texttt{zero-shot} setting, with increase from $0.61$ to $0.86$ AUPRC with simple prompt and all features, and from $0.71$ to $0.79$ with rich prompt and expert features.
\Cref{apd:final-model-chat-history} presents the initial versions of the program, illustrating how the iterative SEDI strategy incrementally refined the model based on feedback.

\begin{figure}[htb!]
    \centering
    \begin{minipage}[t]{0.49\textwidth}
        \centering
        \begin{tcolorbox}[colframe=black!50, colback=gray!10, title=LLM-generated aTRH computable phenotype]
            \lstset{language=Python, 
                basicstyle=\linespread{0.9}\ttfamily\scriptsize, 
                numberstyle=\tiny\color{gray},
                keywordstyle=\color{blue},
                commentstyle=\color{dkgreen},
                stringstyle=\color{mauve},
                showstringspaces=false,
                xleftmargin=-0.25cm,
                xrightmargin=-0.25cm,
                identifierstyle=\color{dkgreen}}
\begin{lstlisting}
def predict_aTRH(*features):
  prob = 0.0
  if high_BP_during_htn_meds_3 >= 2:
    prob += 0.31  # 0.4
  if sum_enc_during_htn_meds_4_plus >=2:
    prob += 0.28  # 0.4
  if mean_diastolic > 80:
    prob += 0.07  # 0.1
  if mean_systolic > 140:
    prob += 0.12  # 0.1
  if high_BP_during_htn_meds_2 > 5:
    prob += -0.16 # 0.1
  if high_BP_during_htn_meds_3 > 5:
    prob += 0.02  # 0.1
    
  prob = min(1.0, prob)

  if Med_Potassium_N > 0 and Dx_HypoK_N > 0:
    prob *= 0.45  # 0.5

  if mean_systolic<130 and mean_diastolic<75:
    prob *= 0.77  # 0.5
        
  return prob
\end{lstlisting}
        \end{tcolorbox}
        \captionof{figure}{
            Final aTRH CP generated by \texttt{gpt-4o+SEDI} strategy with expert features and rich prompts. Constants were obtained using a parameter optimizer. Original constants are shown as comments.
            Held-out performance AUPRC of $0.85$ and AUROC of $0.94$.
        }
        \label{fig:aTRH-final-model-gpt-4o}
    \end{minipage}
    \hfill
    \begin{minipage}[t]{0.49\textwidth}
        \centering
        \begin{tcolorbox}[colframe=black!50, colback=gray!10, title=FEAT aTRH computable phenotype]
            \lstset{language=Python, 
                basicstyle=\linespread{0.9}\ttfamily\scriptsize, 
                numberstyle=\tiny\color{gray},
                keywordstyle=\color{blue},
                commentstyle=\color{dkgreen},
                stringstyle=\color{mauve},
                showstringspaces=false,
                xleftmargin=-0.25cm,
                xrightmargin=-0.25cm,
                identifierstyle=\color{dkgreen}}
\begin{lstlisting}
def predict_aTHR_FEAT(*features):
  acc = 0
  if sum_enc_during_htn_meds_3 > 1:
    acc += 1.33
  if median_enc_during_htn_meds_4_plus >1.25:
    acc += 0.49
  if mean_systolic >= 128.64:
    acc += 0.95
  if max_CALCIUM >= 10.15:
    acc += 0.4
  if re_htn_spec_sum > 40:
    acc += 0.42
    
  acc += -0.52*sd_enc_during_htn_meds_2
  offset = -3.96 
  
  logit = 1/(1+exp(-(offset+acc)))
  
  return logit
\end{lstlisting}
        \end{tcolorbox}
        \captionof{figure}{
            Final aTRH CP generated by FEAT, adapted to Python from Figure 5b of \citet{lacavaFlexibleSymbolicRegression2023}.
            Held-out performance AUPRC of $0.80$ and AUROC of $0.94$.
        }
        \label{fig:aTRH-final-model-feat}
    \end{minipage}
\end{figure}

\section{Discussion}
\label{sec:discussion}

From \cref{fig:ablation}, we can see that the rich phenotype description strongly impacted LLM-derived CP performance for most phenotypes.
This is expected, particularly for the heuristic phenotypes, as providing a precise specification in the prompt narrows the complexity of the task for the LLM. 
If the LLMs could perform well using the much simpler prompt, this would be a much more scalable and generalizable approach. 
Notably, \texttt{gpt-4o} appeared to perform much better than other LLMs using simple prompts, particularly using the iterative strategy and/or with an expert-curated feature set. In fact, from \cref{tab:held-out-test-metrics} we see that the best average performance of \texttt{gpt-4o+SEDI} with simple prompts outperformed its counterpart with rich prompts.

While the rich prompts contain the natural language definitions of the heuristics, they do not provide explicit code. As a result, the LLM is not always able to perfectly replicate the heuristic phenotypes as executable programs.
At the same time, it reflects a real-world scenario in which no expert-driven feature selection is performed.

The SEDI strategy appeared to improve CP performance for many of the underperforming LLM-generated CPs. 
For the most complicated outcome, aTRH by chart review, SEDI appeared to improve all \texttt{gpt-4o-mini} CPs and both \texttt{gpt-4o} CPs provided with only the simple phenotype description. 
In fact, for the most minimal prompt designs (\textit{i.e.}, 
simple prompts using all features), SEDI appeared to improve CP performance for \texttt{gpt-4o} and \texttt{gpt-4o-mini} across all phenotypes, and in cases where it did not, it caused no harm.
However, for several of the outcomes, these SEDI-associated improvements did not yield equivalent performance to that of using the richer prompts. 
We expect future improvements to the SEDI process will lead to further improvements. It is also worth noting that \texttt{gpt-3.5-turbo} only showed substantial improvements with SEDI for the HTN-HypoK and aTRH heuristics.

Across experiments, the impact of providing an expert-curated feature set was inconsistent, when a rich prompt is presented to the model. For \texttt{gpt-4o} and the simple phenotype descriptions, it appeared overall associated with improved model performance. However, there were situations in which it appeared associated with worse performance, including for \texttt{gpt-3.5-turbo} and HTN-HypoK.
The set of expert features had the greatest impact on \texttt{gpt-4o-mini} with simple prompts, suggesting that a smaller search space can compensate for the lack of a detailed phenotype description.

Additionally, the full feature set --- comprising more than $300$ features --- expands the search space and often includes features with overlapping semantics.
Using expert features appears to help by narrowing the search space and reducing prompt length.
However, this is not a general rule, as competitive performance is still observed with the full feature set in several cases.
This suggests that while curated features can reduce complexity, the method remains effective even when such features are not available.
Model capability also plays a significant role --- for example, \texttt{gpt-3.5} consistently underperforms compared to \texttt{gpt-4} variants. 

Anecdotally, we observe that LLMs sometime tried to ``cheat" on the task by, \textit{e.g.}, creating random weights, or importing external libraries to fit a classifier, despite not having access to labels. 
However, the occurrence of these programs is drastically reduced when using SEDI strategy. 
In fact, when using SEDI, we observe that LLMs will, in some instances, create and refine a list of weights for the features, but without SEDI occasionally will generate weights as \texttt{weights=np.random.rand(len(features))}.  
Occurrences of programs with random weights or shortcuts are more common with zero-shot strategies for \texttt{gpt-3.5-turbo} and \texttt{gpt-4o-mini}.

From \cref{fig:pareto-plots} we observe that ML-based strategies (except DT) tend to outperform LLM-based strategies on average. However, the performance differences between the best LLM models (\texttt{gpt-4o+SEDI}) and FEAT are typically small and \texttt{gpt-4o+SEDI} was superior for the HTN-hypoK heuristic.
Notably, $\texttt{gpt-4o+SEDI}$ underperforms on the simplest tasks (HTN Heuristic/Dx) where other ML methods excel, but is competitive on the more complex HTN-HypoK heuristic.
The benefits of the SEDI strategy were apparent for \texttt{gpt-4o-mini} across both HTN and aTRH phenotypes and for \texttt{gpt-4o} for the HTN heuristic.

In \cref{fig:pareto-plots} we see that \texttt{gpt-4o+SEDI} outperforms the other LLM variants, although their performance is close, with overlapping confidence intervals.
\texttt{gpt-4o+SEDI} performance falls between the simplicity of the DT model and the complexity of the RF model, demonstrating a good balance between AUPRC and model size.

Even though \texttt{gpt-4o+SEDI} is outperformed by FEAT in individual trials, selecting the LLM final model from across the cross-validation experiment yields a final LLM model with performance comparable to that of the FEAT model, as reported in \cref{tab:held-out-test-metrics}.
If we consider model size as a proxy for interpretability, we see that \texttt{gpt-4o+SEDI} achieves better results than the black-box model generated by RF, whose ensemble exceeds five thousand nodes. In classification metrics, it achieves an AUPRC of $0.85$, falling short of the best-performing method, RF, which reaches $0.90$.
However, compared to the traditional RF approach, our proposed SEDI method combined with the most recent LLM offers a better balance between performance and interpretability.

We also note from \Cref{tab:held-out-test-metrics} that the zero-shot strategy without a rich prompt and expert features generates a larger model than that with SEDI, yet with decreased performance.
When using a rich prompt and expert features, it generated an oversimplified model that likewise failed to achieve competitive performance.
This indicates that SEDI may play an important in refining models to improve size and performance.
The final CP for the \texttt{gpt-4o} using the zero-shot strategy has low variability, likely due to confidence intervals being computed by bootstrapping performance of a single selected program on a held-out test set.

The final CP model found by \texttt{gpt-4o+SEDI}, depicted in \cref{fig:aTRH-final-model-gpt-4o}, reveals a concise set of intuitive, interpretable rules. We argue that interpretability comes mainly from the fact that the program can be inspected, and all instructions are clear enough to be used in clinical practice.

Taking as an example the final CP model generated by \texttt{gpt-4o+SEDI}, the program clearly assigns cumulative probabilities based on observed conditions.
The program begins by checking for high blood pressure in patients prescribed three hypertension medications and correspondingly increases the cumulative probability for aTRH. The program then continues to adjust the probability by interrogating other features more deeply and incorporating additional information.
After calculating the final cumulative probability, it compensates for potential false negatives by reducing the probability based on number of potassium supplement prescriptions (\texttt{Med\_Potassium\_N}) and hypokalemia diagnoses (\texttt{Dx\_HypoK\_N}), which likely relates to bias in the subject sampling in the training set.

Overall, this final CP is interpretable, intuitively represents predictor interactions, and achieves competitive performance.
Moreover, we demonstrated that supervised parameter optimization has the potential to considerably improve the performance of the LLM-trained models by tuning the parameters of the model, which are close to the actual parameters the LLM generated (represented as comments in the code).

We assess potential biases in the final model generated by \texttt{gpt-4o+SEDI} by evaluating its performance on each subpopulation defined by gender and race.
The final model does not incorporate race or gender, but it shows worse AUPRC performance in Black women and white women. A detailed analysis is presented in \cref{apd:bias}.

\paragraph{Limitations} We note several limitations. 
First, we focused exclusively on hypertension and subphenotypes thereof; it is possible that different methods or settings would excel when applied to different phenotypes.
Second, we did not evaluate the performance of other, potentially more advanced LLMs like gpt-o1, Claude or llama.
Third, there may be variations of the SEDI strategy that would provide more effective feedback to the LLMs; a full experimental study of these many options is beyond the current scope of this paper.
The generated models are not calibrated to mitigate bias across subpopulations, which could potentially be addressed by incorporating fairness objectives in the prompts.
Finally, our experiment is confined to a limited set of patients from a single (albeit large, multi-site) health system, partially due to the intensive nature of chart review. Because of this we cannot rule-out differences in the performance of these CPs across distinct patient populations. 

\section{Conclusions and Future Work}
\label{sec:conclusions}

State-of-the-art LLMs generate reasonably accurate and concise CPs for hypertension phenotypes, even when a simple prompt is presented.
When given detailed and focused prompts and equipped with data-driven, iterative feedback (\textit{i.e.}, SEDI), LLM-generated CPs are competitive with those trained using supervised ML.
We observe in general that traditional supervised ML approaches leveraging chart-reviewed examples still outperform LLM-derived CPs, but yield models that are much larger and require access to a much larger set of expert-labeled data.

We have also established the potential for LLMs to iteratively learn CPs expressed as intelligible Python code.

Our work presented a novel approach for using LLMs to generate CPs automatically, aiming to reduce manual feature engineering and leaning towards scalable solutions.
We have made our SEDI framework publicly available so that it can be readily adapted to developing CPs for other conditions (\textit{e.g.}, diabetes complications), given a small set of training examples.

Future work could extend the SEDI-based LLM approach used here, consider an expanded set of phenotyping tasks, or look at the performance of an iterative computable phenotyping system in prospective clinical settings. 

\paragraph*{Data and Code Availability}
Study data was extracted from the University of Pennsylvania Healthcare System (UPHS) electronic health record and cannot be shared publicly to protect the privacy of the subjects. 
However, it can be shared upon request and subject to relevant approvals.
All our methods, experiments, and post-processing analysis are publicly available at \url{https://github.com/cavalab/htn-phenotyping-with-llms}.

\paragraph*{Institutional Review Board (IRB)}
This study was reviewed and approved by University of Pennsylvania Institutional Review Board (\#827260), which approved a waiver of informed consent. 
This study followed all relevant ethical regulations. 

\acks{%
This work was supported by Patient-Centered Outcomes Research Institute (PCORI) Award (ME-2020C1D-19393). W.G. La Cava was supported by National Institutes of Health (NIH) grant R01-LM014300. 
G.S.I.A.~is supported by Coordena\c{c}\~{a}o de Aperfei\c{c}oamento de Pessoal de N\'{i}vel Superior (CAPES) finance Code 001.
The statements in this work are solely the responsibility of the authors and do not necessarily represent the views of PCORI, its Board of Governors or Methodology Committee, or the NIH.
}

\bibliography{refs,CHIL-LLM-CP-do-not-edit}

\appendix

\section{Prompts for the SEDI strategy}\label{apd:sedi-prompts}

In this section, we provide a detailed description of the prompts used in the synthesize-execute-debug-inspect (SEDI) strategy.

The iteration begins with the same message for both SEDI and the zero-shot strategy, which includes a description of the phenotype and the available features. Then, after receiving the response from the LLM, we parse it to extract the generated program and evaluate it for execution errors.

If execution errors occur, the following message is used as a prompt:

\vspace{5pt}\hrule
\paragraph{Debug prompt} \hl{Python encountered an error when trying to execute the function. Error Message: \textless{}error traceback as formatted text\textgreater{}. Please try again. **MAKE ABSOLUTELY SURE TO RETURN A SYNTACTICALLY VALID PYTHON FUNCTION.}
\vspace{5pt}\hrule\vspace{5pt}

If the program executes without errors, we evaluate its performance by checking for false positives and false negatives. For each case, we randomly select up to $10$ examples from the training data and format them as input dictionaries.
Notice that, in the worst case scenario, the model will see only $20$ training samples each iteration, meaning that the training uses less data than a supervised ML method.

The updated instruction prompt consists of three components: the \textit{performance report} message, the \textit{false positive examples} (if applicable), and the \textit{false negative examples} (if applicable).
The second and third components are used only if there is any false positive or negative case, otherwise the model does not receive specific examples.
During the first iteration, no improvement message is provided to the LLM. After the first iteration, the performance report message includes either a \textit{reinforcement} message if the changes improve the model or a negative message otherwise.

\vspace{5pt}\hrule
\paragraph{Performance report message} {
    \parindent0pt
    \hl{We evaluated the prediction function you provided on a set of \textless{} training dataset size\textgreater{} patients. \textless{}reinforcement message, if applicable\textgreater{}. Using the performance feedback below, please refine the Python function. 
        
    \# Overall Performance
    
    Area Under the Receiver-Operating Curve (AUROC): \textless{}AUROC, with 3 decimal places\textgreater{}
    
    Area under the precision-recall curve (AUPRC): \textless{}AUPCR, with 3 decimal places\textgreater{}
    
    The False Positive Rate is \textless{}FP rate, as percentage\textgreater{}
    
    The False Negative Rate is \textless{}FN rate, as percentage\textgreater{}
    }
}
\vspace{5pt}\hrule\vspace{5pt}

\vspace{5pt}\hrule
\paragraph{False positive examples} {
    \parindent0pt
    \hl{\# Analysis of False Positives

    Please refine the function so that the \textless{}number of false positives\textgreater{} False Positives have lower predicted probabilities
    
    Below you will find \textless{}number of FP examples\textgreater{} example patients with false positive assessments to assist you in prescribing changes to the predict\_hypertension function:
    
    \textless{}List containing the FP examples, formatted as dictionaries\textgreater{}
    }
}
\vspace{5pt}\hrule\vspace{5pt}

\vspace{5pt}\hrule
\paragraph{False negative examples} {
    \parindent0pt
    \hl{\# Analysis of False Negatives
        
    Please refine the function so that the \textless{}number of false negatives\textgreater{} False Negatives have higher predicted probabilities
    
    Below you will find \textless{}number of FN examples\textgreater{} example patients with false negative assessments to assist you in prescribing changes to the predict\_hypertension function:
    
    \textless{}List containing the FN examples, formatted as dictionaries\textgreater{}
    }
}
\vspace{5pt}\hrule\vspace{5pt}

Finally, the prompt concludes with a \textit{summary} message that reiterates the task description.

\vspace{5pt}\hrule
\paragraph{Summary} {
    \parindent0pt
    \hl{
    \# Summary of Request
    
    Please create an updated Python function named `predict\_hypertension` that achieves fewer false positives and fewer false negatives than the one you previously provided.
    
    The function should assess whether each patient (represented as rows) has evidence of \textless{}phenotype\textgreater{}.
    
    As before, the function takes a pandas DataFrame named `df` as input.
    
    Recall that the available columns and their meanings are provided as key value pairs in a dictionary previously provided.
    
    As before, you may only use the features whose names appear in this dictionary.
    
    As before, your response must contain only a Python function, with no comments or explanations, that strictly follows the given description.
    }
}
\vspace{5pt}\hrule\vspace{5pt}

\section{Feature construction}\label{apd:all-feature-description}

As previously described, $331$ features were extracted from the EHR.
Demographic and encounter features included age, race, sex, binned distance from zip code 19104, weight, BMI, blood pressure, and the number of encounters with elevated blood pressure (systolic/diastolic BP  $>= 140/90$ mmHg). Longitudinal features were aggregated as minimum, maximum, median, standard deviation, and skewness.

The $34$ most common ambulatory laboratory test results (\textit{i.e.}, complete metabolic panel, complete blood count with differential, lipids, thyroid stimulating hormone, and hemoglobin A1c) with missingness in fewer than one-third of subjects were summarized as minimum, maximum, median, 1st quartile, and 3rd quartile. Diagnosis codes for hypertension, associated comorbidities, and other indications for anti-hypertensive medications were aggregated and summarized as median per year and sum. Medication prescriptions were summarized as both the number of days prescribed for each antihypertensive class and the count of encounters while prescribed $1$, $2$, $3$, or $4$ or more anti-hypertensive medications, described as sum, median, standard deviation, and skewness, as well as the sum of encounters with elevated blood pressure. Regular expressions were applied to clinical notes to identify mentions of `hypertension' and variants thereof, summarized as counts.

Features with values outside of physiologically reasonable ranges, with fewer than $5\%$ non-zero counts, or with variance $<0.05$ were excluded. Missing values were median imputed.

In the original paper from which the data were extracted \citep{lacavaLearningConciseRepresentations2019}, the diagnoses were manually ascertained through chart review.
A study physician (a MD graduate with experience as a medical officer) reviewed clinical charts and classified subjects according to three phenotypes. Classification was based on the JNC7 Guidelines on Prevention, Detection, Evaluation, and Treatment of High Blood Pressure \citep{chobanian2003seventh}.
The heuristics were proposed by expert clinicians and iteratively refined using the training data to assess how many cases were labeled as true positives.
Unclear cases were reviewed by one additional study physician. The chart review was documented in a single Redcap form, with the reviewer’s conclusion and the underlying evidence.

\cref{tab:all-features} summarizes all features present in the dataset.
All features can be considered potential risk factors for different hypertension phenotypes.
Some features represent descriptive statistics of lab tests, with a suffix indicating the corresponding statistic. When multiple statistics are available, they are separated by a forward slash in the feature name.

\begin{table}[!ht]
    \caption{All features in the dataset. Features with multiple variations (indicated by words separated by a forward slash) mean that all options exist as features}
    \label{tab:all-features}
    \centering
    \footnotesize\setlength{\tabcolsep}{3pt}
    \renewcommand{\arraystretch}{1.1}
    \begin{tabular}{@{}llp{7.25cm}@{}}
    \toprule\midrule
    Group & Feature & Feature Description \\ \midrule
    \multirow{6}{*}{Demo} & Age& Patient’s age at right-censoring date \\ \cline{2-3}
     & Sex& Patient’s documented Sex (1 = male, 0 = female) \\ \cline{2-3}
     & Race (Black)& 1 = black, 0 = non-black \\ \cline{2-3}
     & Race (Other)& \begin{tabular}[c]{@{}l@{}}1 = Asian, other, mixed, Native American, \\ Pacific Islander,\\ 0 = black or white\end{tabular} \\ \cline{2-3}
     & Race (White)& 1 = white, 0 = non-white \\ \cline{2-3}
     & ZIP\_CAT & Distance from patient’s home to 19104, binned \\ \hline
    \multirow{2}{*}{Encounter} & Practice Site& \begin{tabular}[c]{@{}l@{}}Code for healthcare site (not one-hot encoded),\\ common service for all source systems.\end{tabular} \\ \cline{2-3}
     & Practice type& 1 = internal medicine practice, 0 = family medicine practice\\ \hline
    \multirow{2}{*}{BMI/Weight} & weight\_\{statistic\} & Min/max/median/sd/skewness of weights \\ \cline{2-3}
     & bmi\_\{statistic\} & Min/max/median/sd/skewness of BMI \\ \hline
    \multirow{7}{*}{BP} & bp\_n & Total number of blood pressure (BP) measurements \\ \cline{2-3}
     & \{statistic\}\_systolic/diastolic & Max/mean/median/sd/skew of systolic/diastolic BP measured \\ \cline{2-3}
     & high\_bp\_n & Number of high blood pressure cases (SBP \textgreater{}= 140 or DBP \textgreater{}= 90) \\ \cline{2-3}
     & \{statistic\}\_high\_bp\_systolic/diastolic & \begin{tabular}[c]{@{}l@{}} Mean/median/sd/skew of systolic/diastolic BP of \\ all high blood pressure measurements\\ (SBP \textgreater{}=140 or DBP \textgreater{}= 90)\end{tabular} \\ \cline{2-3}
     & \{statistic\}\_high\_bp\_n\_yr & \begin{tabular}[c]{@{}l@{}} Median/sd/skew of systolic/diastolic BP \\ for high blood pressure measurements\end{tabular} \\ \hline
    \multirow{6}{*}{Labs Dx} & \{statistic\}.lab\_XXX & Maximum/minimum/median/q1/q3 of XXX lab test \\ \cline{2-3}
     & \{statistic\}\_ICD\_XXX (Dx) & Median/sum XXX ICD-9 and ICD-10 codes, by year \\ \cline{2-3}
     & \{statistic\}\_XXX (disease name) & Median/sum XXX disease name, by year \\ \cline{2-3}
     & Dx\_N & Number of total ICD-9 and ICD-10 codes (PX\_DX\_ID) \\ \cline{2-3}
     & enc\_N & Number of OUTPATIENT (including INFUSION VISIT) encounters \\ \cline{2-3}
     & dx\_days\_x & Days from 1st Dx to last Dx in system \\ \hline
      \multirow{7}{*}{Medication} & HTN\_MED\_days\_XXX & Days on med XXX (including anti-HTN and Potassium Supplement) \\ \cline{2-3}
     & MED\_N & Number of medication prescriptions total \\ \cline{2-3}
     & high\_bp\_during\_htn\_meds\_\{meds\} & Number of high BP measurements during 1/2/3/4+ anti-HTN meds \\ \cline{2-3}
     & \{statistic\}\_enc\_during\_htn\_meds\_\{meds\} & Number of OUTPATIENT encounters during 1/2/3/4+ meds \\ \midrule\bottomrule
    \end{tabular}
    Continued on next page
\end{table}

\begin{table}[!ht]
    \ContinuedFloat
    \caption{All features in the dataset. Features with multiple variations (indicated by words separated by a forward slash) mean that all options exist as features (Continued)}
    \label{tab:all-features:2}
    \centering
    \footnotesize\setlength{\tabcolsep}{3pt}
    \renewcommand{\arraystretch}{1.1}
    \begin{tabular}{@{}llp{9.5cm}@{}}
    \toprule\midrule
    Group & Feature & Feature Description \\ \midrule
    \multirow{5}{*}{Medication} & N\_med\_k\_chlo\_enc & \begin{tabular}[c]{@{}l@{}}Number of encounters on\\ POTASSIUM\_CHLORIDE/POTASSIUM\_GLUCONATE\end{tabular} \\ \cline{2-3}
     & sd\_med\_k\_chlo\_enc & \begin{tabular}[c]{@{}l@{}}Standard deviation of number (by year) of encounters on\\ POTASSIUM\_CHLORIDE/POTASSIUM\_GLUCONATE\end{tabular} \\ \cline{2-3}
     & skewness\_med\_k\_chlo\_enc & \begin{tabular}[c]{@{}l@{}}Skewness of number (by year) of encounters on\\ POTASSIUM\_CHLORIDE/POTASSIUM\_GLUCONATE\end{tabular} \\ \hline
    \multirow{4}{*}{\makecell{Heuristic\\Features}} & low\_K\_N & Number of low potassium test results \\ \cline{2-3}
     & test\_K\_N & Number of potassium test results \\ \cline{2-3}
     & Med\_Potassium\_N & Number of potassium supplement medication subscriptions \\ \cline{2-3}
     & Dx\_HypoK\_N & Number of Hypokalemia Dx \\ \hline 
    \multirow{6}{*}{\makecell{HTN Score\\Features}} & ICD\_hyp\_sum & HTN ICD codes \\ \cline{2-3}
     & Med\_HTN\_N & Anti-HTN meds prescriptions \\ \cline{2-3}
     & bp\_hyp\_norm & high\_bp\_n/bp\_n \\ \cline{2-3}
     & icd\_hyp\_sum\_norm & ICD\_hyp\_sum/Dx\_N \\ \cline{2-3}
     & MED\_HTN\_N\_norm & MED\_HTN\_N/MED\_N \\ \cline{2-3}
     & re\_hyp\_spec\_norm & re\_hyp\_spec/words\_n \\ \hline
    \multirow{7}{*}{Regex} & re\_htn\_\{statistic\} & \begin{tabular}[c]{@{}l@{}}Sum/max/mean/median/sd/skewness of regex counts in clinical\\notes for hypertension\end{tabular} \\ \cline{2-3}
     & re\_htn\_spec\_\{statistic\} & \begin{tabular}[c]{@{}l@{}}Sum/max/mean/median/sd/skewness of regex counts in clinical\\notes for hypertension (specific, excluding preliminary negations)\end{tabular} \\ \cline{2-3}
     & re\_htn\_teixeira\_\{statistic\} & \begin{tabular}[c]{@{}l@{}}Sum/max/mean/median/sd/skewness of regex counts in clinical\\notes for hypertension (regex used in Teixeira paper)\end{tabular} \\ \cline{2-3}
     & re\_word\_count\_\{statistic\} & Sum word counts in clinical notes \\ \midrule\bottomrule
    \end{tabular}
\end{table}

\section{ML hyper-parameters}\label{apd:hyperparameters}

\cref{tab:other-ml-methods} presents the machine learning methods compared in the second set of experiments.
The hyperparameters for each method were optimized using a gridsearch with a stratified $5$-fold cross-validation.
For each method, the table lists the possible values considered during the gridsearch process to identify the optimal configuration.

\begin{table*}[htb]
\footnotesize
    \caption{Machine learning methods compared in the second set of experiments, with hyper-parameters optimized via gridsearch in a stratified $5$-fold cross-validation. Each list represents the possible values set to optimize using gridsearch.}
    \label{tab:other-ml-methods}
    \centering
    \begin{tabular}{ll}
    \toprule\midrule
    Method & Parameter \\ \midrule
    Decision Tree (DT) & \begin{tabular}[c]{@{}l@{}}max\_features = {[}`auto', `sqrt'{]}\\ max\_depth = {[}10, 20, 30, 40, 50, 60, 70, 80, 90, 100, 110{]}\\ min\_samples\_split = {[}2, 5, 10{]}\\ min\_samples\_leaf = {[}1, 2, 4{]}\end{tabular} \\ \midrule
    \begin{tabular}[c]{@{}l@{}}Logistic Regression\\ with L1 norm (LR L1)\end{tabular} & \begin{tabular}[c]{@{}l@{}}Cs = {[}1.e-06, 1.e-05, 1.e-04, 1.e-03, \\ 1.e-02, 1.e-01, 1.e+00, 1.e+01, 1.e+02, 1.e+03{]}\\ penalty={[}`l1'{]}\\ solver = {[}`liblinear'{]}\end{tabular} \\ \midrule
    Random Forests (RF) & \begin{tabular}[c]{@{}l@{}}n\_estimators = {[}100, 500, 900, 1300, 1700, 2100{]}\\ max\_features = {[}`auto', `sqrt'{]}\\ max\_depth = {[}10, 30, 50, 70, 90, 110{]}\end{tabular} \\ \midrule
    \begin{tabular}[c]{@{}l@{}}Feature Engineering\\ Automation Tool (FEAT)\end{tabular} & \begin{tabular}[c]{@{}l@{}}max\_depth=[6]                                             \\ max\_dim = [10]\\ objectives= [{[}``fitness",``size"{]}]\\ sel=[`lexicase']\\ gens = [200]\\ pop\_size = [1000]\\ stagewise\_xo = [True]\\ scorer=[`log']\\ ml=[`LR']\\ fb=[0.5]\\  classification=[True]\\ functions= {[}``split",``and",``or",``not",``b2f"{]}\\ split=[0.8]\\ normalize=[False]\\ corr\_delete\_mutate=[True],\\ simplify=[0.005]\end{tabular} \\ \midrule\bottomrule
    \end{tabular}
\end{table*}

\section{Boxplot visualization for the settings study}\label{apd:boxplot}

\begin{figure}[htb]
    \centering
    \includegraphics[width=\linewidth]{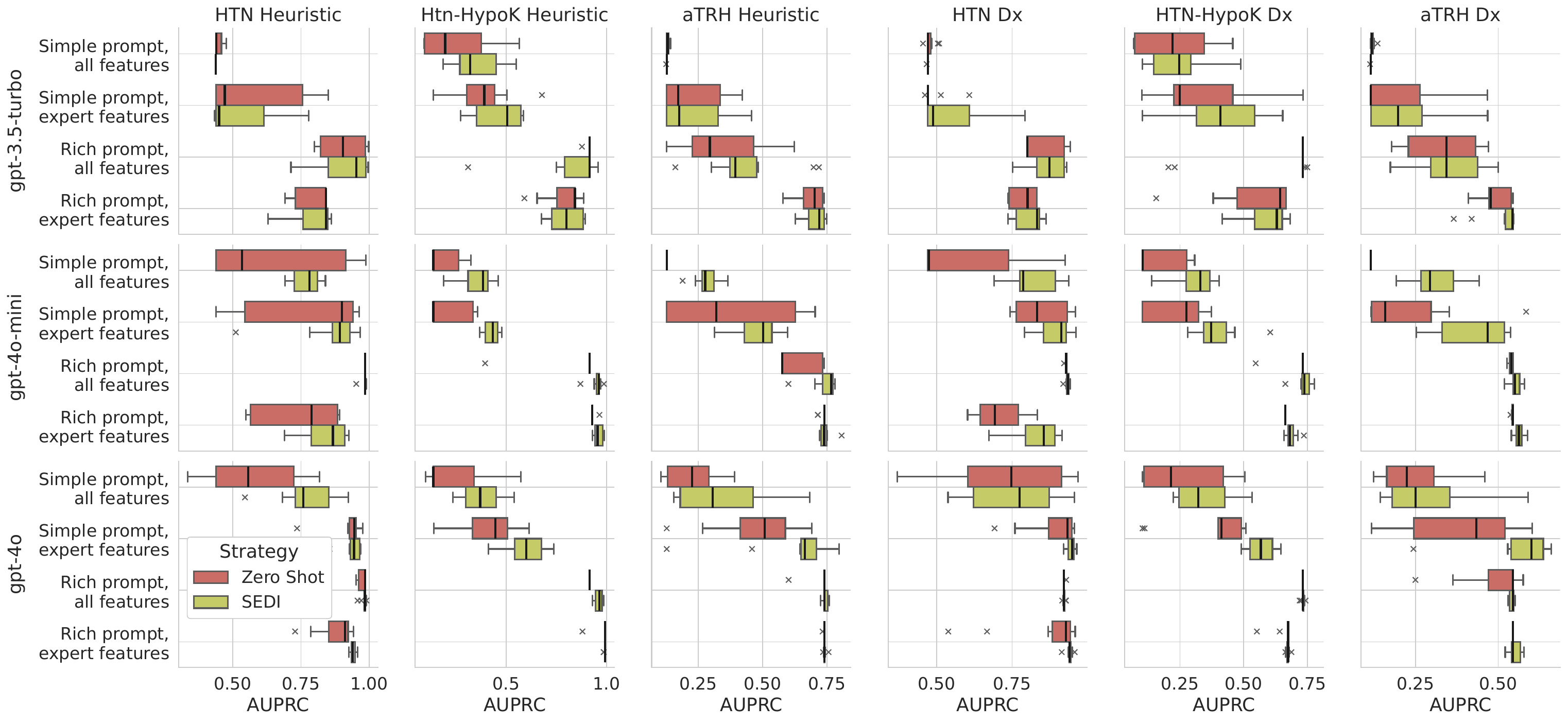}
    \caption{Average AUPRC on each held-out fold of different LLM-generated CPs, compared to other interpretable machine learning methods.
        LLMs results are shown using rich prompts and expert
        features, with and without SEDI training.}
    \label{fig:apd-boxplot}
\end{figure}

\cref{fig:apd-boxplot} reports the same data from the settings study (\Cref{fig:ablation}), using boxplots instead of barplots.
While the barplot provides a cleaner view of average performance, making it easier to compare two methods, the boxplot gives us distribution estimates.

\section{AUPRC comparison for aTRH}\label{apd:auprc-all-methods}

\begin{figure*}[htb]
    \centering
    \includegraphics[width=\linewidth]{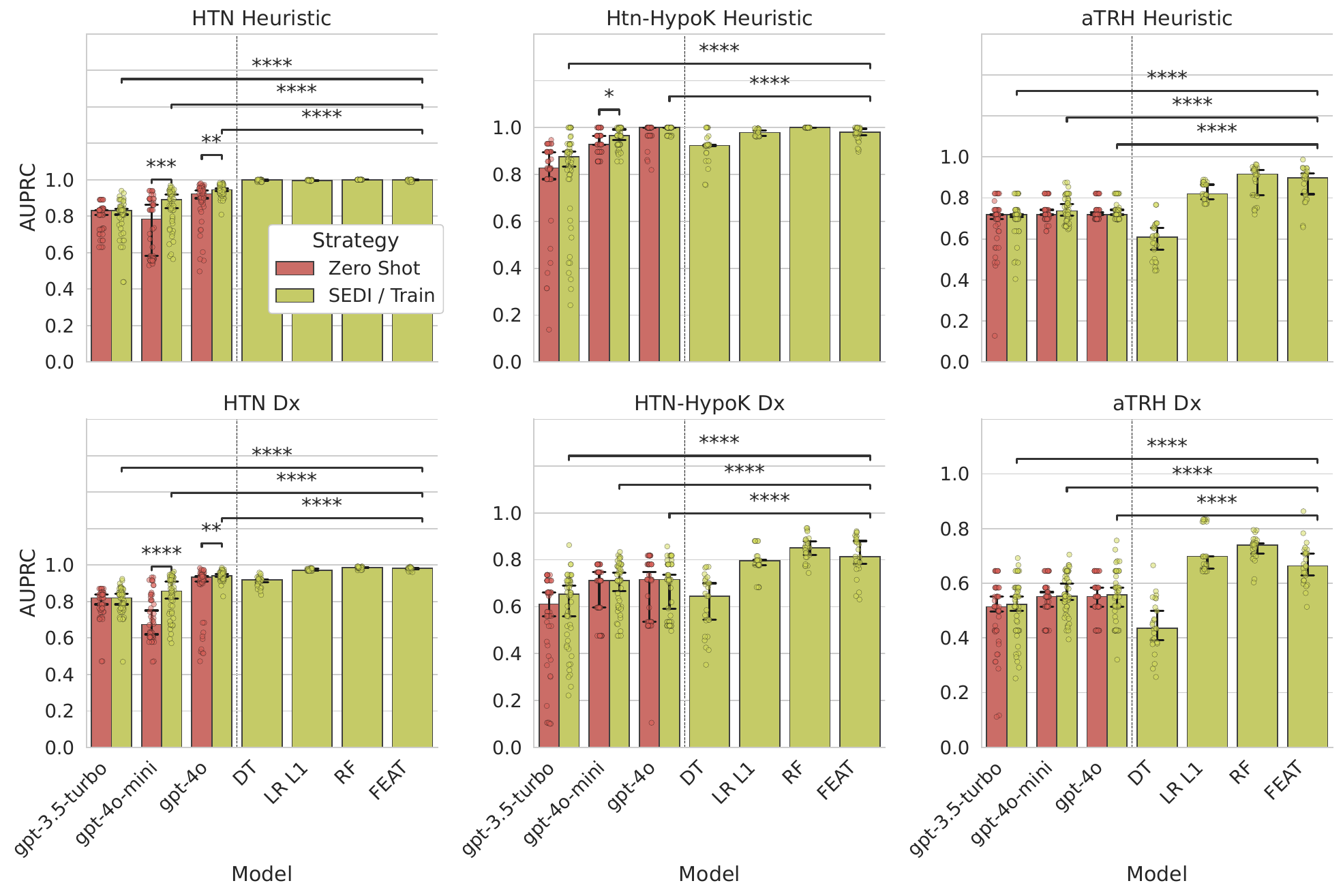}
    \caption{
        Average AUPRC cross-validation of different LLM-generated CPs, compared to other interpretable machine learning methods from~\cite{lacavaFlexibleSymbolicRegression2023}. 
        LLMs results are shown using rich prompts and expert 
        features, with and without SEDI training.
        Error bars depict the $95\%$ confidence interval across different seeds. In addition to the bars, the performance of each run is depicted as a dot overlaying the bars.
        Statistical comparisons are conducted using Mann-Whitney-Wilcoxon two-sided test with Holm-Bonferroni corrections. 
       *: 1.00e-02 $<$ p $\leq$ 5.00e-02; **: 1.00e-03 $<$ p $\leq$ 1.00e-02 ***: 1.00e-04 $<$ p $\leq$ 1.00e-03 ****: p $\leq$ 1.00e-04.
        \Cref{apd:boxplot} has a boxplot visualization of the same data.
    }
    \label{fig:apd-auprc-all-methods}
\end{figure*}

\cref{fig:apd-auprc-all-methods} reports a comparison of LLM-generated CPs held-out validation performance in $5$-fold cross-validation, comparing the LLMs to conventional interpretable ML methods DT and LR L1, a representative non-interpretable ML method RF, and a representative symbolic regression method FEAT.  
We observe that the SEDI approach appeared associated higher performance for some LLM, phenotype combinations, particularly for \texttt{gpt-4o-mini} for the more complicated aTRH phenotypes. 
The average performance of CPs generated by \texttt{gpt-4o} for HTN-HypoK heuristic outperformed FEAT, however in all other cases the LLM-generated CPs showed slightly worse average performance.
In despite of that, particular runs generated good performing CPs.

\section{Using optimizers to fine-tune the final model}\label{apd:final-model-tuning}

We performed supervised parameter optimization on the final \texttt{gpt-4o+SEDI} model using a black-box optimizer nevergrad \citep{nevergrad}.
The optimizer aimed to maximize AUPRC by exploring different combinations of hyperparameters in the training data.
The approach was implemented by manually adjusting the python function to include all numeric parameters as arguments and then using an gradient-free, adaptative optimization algorithm, to iteratively evaluate different combinations of values.
The process can be applied to our generated methods as it does not rely on prior knowledge of the model's internal structure, but we also notice that it has limitations depending on the initial starting point.

In this proof-of-concept, we demonstrate that LLM-constructed models can be considerably refined by leveraging available outcome data, searching for sub-optimal parameter configurations that further improves the performance of our generated method.
The downside of this approach is that it requires a lot of outcome data in order to perform a parameter optimization of the generated program in a supervised learning fashion.

\section{Dialog for generating the final aTRH model}\label{apd:final-model-chat-history}

\figureref{fig:iterations} displays the first three generated programs before the final aTRH model was obtained by \texttt{gpt-4o+SEDI} strategy.
Unlike the reported model in \cref{sec:results}, the outputs here have not been processed for clarity.
We show the first three programs just for illustrative purposes.

The \texttt{Initial model} was evaluated and achieved an AUPRC of $0.54$, AUROC of $0.91$, a FP rate of $8.0\%$ and FN rate of $9.8\%$.
The feedback prompt provided $10$ FP and FN examples.

The LLM then presented the \texttt{Model after first iteration of SEDI}, which included more features and if-else statements, but kept the core instructions from the previous model. The metrics changed to $0.56$ AUPRC, $0.91$ AUROC, $1.1\%$ FP, and $84.1\%$ FN.
We can see that the LLM performed some adjustments, improving some metrics but presenting an extremely high FN rate. The prompt states to create a new model that achieves fewer false positive and fewer false negatives.

The next output updated the model with a new set of conditionals, reported as \texttt{Model after second iteration of SEDI}, while still keeping the core of the previous programs. It was evaluated to $0.60$ AUPRC, $0.93$ AUROC, $4.2\%$ FP, and $40.2\%$ FN.

\begin{figure*}[htb]
    \centering
    \begin{tcolorbox}[colframe=black!50, colback=gray!10, title=Iterative program refinement using the SEDI strategy]
     \lstset{language=Python, 
        basicstyle=\ttfamily\scriptsize, 
        numberstyle=\tiny\color{gray},
        keywordstyle=\color{blue},
        commentstyle=\color{dkgreen},
        stringstyle=\color{mauve},
        showstringspaces=false,
        identifierstyle=\color{dkgreen}}
\begin{lstlisting}
# Initial model
def predict_hypertension(df: pd.DataFrame) -> np.ndarray:
    probabilities = []

    for _, row in df.iterrows():
        if row['high_BP_during_htn_meds_3'] >= 2 or row['sum_enc_during_htn_meds_4_plus'] >= 2:
            probabilities.append(1.0)
        else:
            probabilities.append(0.0)

    return np.array(probabilities)

# Model after first iteration of SEDI
def predict_hypertension(df: pd.DataFrame) -> np.ndarray:
    probabilities = []

    for _, row in df.iterrows():
        prob = 0.0
        if row['high_BP_during_htn_meds_3'] >= 2:
            prob += 0.5
        if row['sum_enc_during_htn_meds_4_plus'] >= 2:
            prob += 0.5
        if row['high_BP_during_htn_meds_3'] >= 2 and row['sum_enc_during_htn_meds_4_plus'] >= 2:
            prob = 1.0
        if row['mean_diastolic'] < 75 or row['Med_Potassium_N'] == 0:
            prob *= 0.5
        probabilities.append(prob)

    return np.array(probabilities)

# Model after second iteration of SEDI
def predict_hypertension(df: pd.DataFrame) -> np.ndarray:
    probabilities = []

    for _, row in df.iterrows():
        prob = 0.0
        if row['high_BP_during_htn_meds_3'] >= 2:
            prob += 0.4
        if row['sum_enc_during_htn_meds_4_plus'] >= 2:
            prob += 0.4
        if row['mean_diastolic'] > 80:
            prob += 0.1
        if row['mean_systolic'] > 140:
            prob += 0.1
        if row['high_BP_during_htn_meds_2'] > 5:
            prob += 0.1
        prob = min(1.0, prob)
        if row['Med_Potassium_N'] > 0 and row['Dx_HypoK_N'] > 0:
            prob *= 0.5
        probabilities.append(prob)

    return np.array(probabilities)
\end{lstlisting}
    \end{tcolorbox}
    \caption{Initial programs generated by GPT-4o with the SEDI strategy.}
    \label{fig:iterations}
\end{figure*}

\section{Performance across different subgroups}\label{apd:bias}

We assess potential biases in the final model generated by \texttt{gpt-4o+SEDI} by evaluating its performance across different subgroups within the test dataset.
We considered bias relative to subject's documented gender and race.
These variables were selected because they are commonly studied in fairness assessments in machine learning models. 

To evaluate bias, we report how the final model performs on each subpopulation defined by gender and race, reporting classification metrics for each subpopulation and comparing them to the overall performance of the model.
This allows us to determine if the model exhibits disproportionate error rates or other disparities across different demographic groups.
\cref{tab:bias} reports the metrics for the intersecting subpopulations when using the final model generated by the \texttt{gpt-4o+SEDI} strategy, using the rich prompt and expert features set, evaluated for aTRH diagnosis.
\cref{tab:case_sample_counts} reports case counts and sample counts.

\begin{table}[htb]
    \caption{Average AUPRC and AUROC on the held-out test partition on each subpopulation, with race and gender as the distinguishing variables, evaluated for aTRH diagnosis.}
    \label{tab:bias}
    \centering
    \footnotesize
    \begin{tabular}{lccc}
        \toprule\midrule
        Race& Gender & AUPRC &  AUROC  \\
        \midrule
        \multirow[c]{2}{*}{Black} & Female & 0.822 & 0.929 \\
                                  & Male & 0.862 & 0.917 \\ \midrule
        \multirow[c]{2}{*}{White} & Female & 0.824 & 0.966 \\
                                  & Male & 0.903 & 0.981 \\ \midrule
        \multirow[c]{2}{*}{Other} & Female & 0.456 & 0.965 \\
                                  & Male & 0.329 & 0.823 \\
        \midrule\bottomrule
    \end{tabular}
\end{table}

We see differences in AUPRC and AUROC across race and gender. The model's discriminative performance appears highest for White men and lowest for Black women and white women. Other ethnic groups shows a low AUPRC, and prevalence is also low for these subgroups.

\begin{table}[htb]
    \caption{Case counts and sample counts for each subpopulation, with race and gender as the distinguishing variables, evaluated for aTRH diagnosis.}
    \label{tab:case_sample_counts}
    \centering
    \footnotesize
    \begin{tabular}{lcccc}
        \toprule\midrule
        Race& Gender & \# Cases & \# Samples & Prevalence \\
        \midrule
        \multirow[c]{2}{*}{Black} & Female & 26 & 70 & 0.37 \\
         & Male & 13 & 37 & 0.35 \\ \midrule
        \multirow[c]{2}{*}{White} & Female & 17 & 97 & 0.17 \\
         & Male & 15 & 69 & 0.21 \\ \midrule
        \multirow[c]{2}{*}{Other} & Female & 1 & 16 & 0.06 \\
         & Male & 1 & 11 & 0.09 \\
        \midrule\bottomrule
    \end{tabular}
\end{table}

There are several explanations for this apparent bias. LLMs are known to recapitulate biases from their training corpuses. However, for this application a likely explanation is bias in training and testing data. 
While the final model does not directly incorporate race or gender, there are associations between these factors and incorporated features.  
As shown in \cref{tab:bias,tab:case_sample_counts}, Black patients are more likely to be documented as having apparent treatment-resistent hypertension. 
Differences in AUPRC measures are explained in part by the sensitivity of this metric to differences in case prevalence. 
Some of the differences in AUROC are also due to have single case counts/few patients within subgroups (\textit{e.g.}, the `Other' category).
In addition, there are many differences in how women and men interact with the healthcare system, which were not considered in the upstream feature engineering.
These findings warrant follow-up to compare for bias in the ML methods and to mitigate bias in the LLM-based approach.  Approaches for mitigating bias could include accounting for subject factors when samping for FP and FN subjects in the SEDI approach.

In addition, prompts could use classification metrics evaluated across subpopulations and ask the LLM to explicitly consider mitigating bias.
\end{document}